\documentclass[11pt]{article}
\usepackage[]{acl}
\usepackage{times}
\usepackage{todonotes}
\usepackage{booktabs}
\usepackage{latexsym}
\usepackage[T1]{fontenc}
\usepackage{microtype}
\usepackage{CJK}
\usepackage{graphicx}
\usepackage{linguex}
\usepackage{dblfloatfix}
\usepackage{graphicx}
\usepackage{multirow}

\usepackage{microtype}

\title{Probing BERT's priors with serial reproduction chains}

\author{Takateru Yamakoshi$^{1,2}$, Thomas L. Griffiths$^1$,  Robert D. Hawkins$^{1}$ \\
$^1$Princeton University, $^2$The University of Tokyo \\
\texttt{\{takateru,tomg,rdhawkins\}@princeton.edu} }

\date{}

\begin{document}
\maketitle
\begin{abstract}

% motivate the problem...
%We can learn as much about language models from what they \emph{say} as we learn from their performance on targeted benchmarks.
Sampling is a promising bottom-up method for exposing what generative models have learned about language, but it remains unclear how to generate representative samples from popular masked language models (MLMs) like BERT.
The MLM objective yields a dependency network with no guarantee of consistent conditional distributions, posing a problem for naive approaches.
Drawing from theories of iterated learning in cognitive science, we explore the use of \emph{serial reproduction chains} to sample from BERT's priors.
% comparison
In particular, we observe that a unique and consistent estimator of the ground-truth joint distribution is given by a Generative Stochastic Network (GSN) sampler, which randomly selects which token to mask and reconstruct on each step.
We show that the lexical and syntactic statistics of sentences from GSN chains closely match the ground-truth corpus distribution and perform better than other methods in a large corpus of naturalness judgments. 
Our findings establish a firmer theoretical foundation for bottom-up probing and highlight richer deviations from human priors\footnote{Code and data are available at \url{https://github.com/taka-yamakoshi/TelephoneGame}}.
\end{abstract}

\section{Introduction}

Large neural language models have become the representational backbone of natural language processing. 
By learning to predict words from their context, these models have induced surprisingly human-like linguistic knowledge, from syntactic structure \cite{linzen2020syntactic, tenney2019bert,warstadt2019neural} and subtle lexical biases \cite{hawkins2020investigating} to more insidious social biases and stereotypes \cite{caliskan2017semantics,garg2018word}.
At the same time, efforts to probe these models have revealed significant deviations from natural language \cite{braverman2020calibration,holtzman2019curious,dasgupta2020analyzing}.
Observations of incoherent or ``weird'' behavior may often be amusing, as when a generated recipe begins with ``1/4 pounds of bones or fresh bread'' \cite{shane2019you}, but also pose significant dangers in real-world settings \cite{bender2021dangers}. 

\begin{figure}[t!]
\begin{center}
\includegraphics[width=0.99\linewidth]{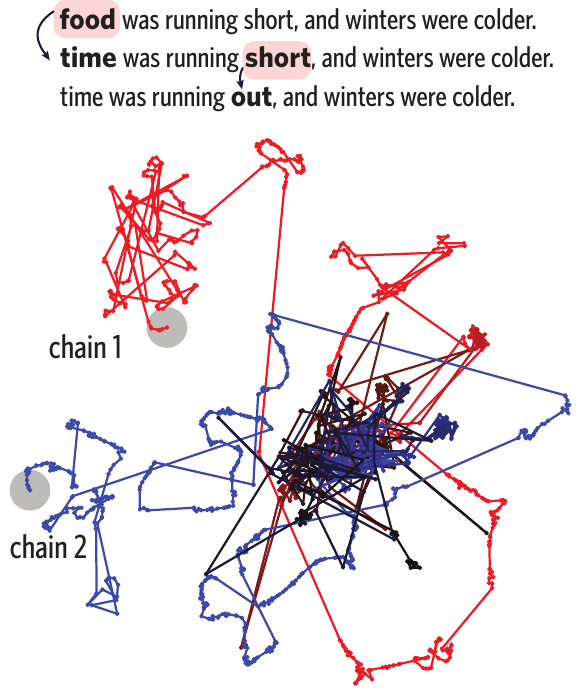}
\caption{We use a \emph{serial reproduction} method to probe BERT's prior over possible sentences (visualization of reproduction chains obtained by running t-sne on sentence embeddings; chains are color-coded and fade to black across their burn-in period).}
\label{fig:front}
\end{center}
\end{figure}

\begin{table*}[ht]
    \centering
    %\footnotesize
    \begin{tabular}{l|l|p{280pt}}
      \multicolumn{2}{c|}{Type of unnaturalness} & Example \\
      \hline
      \hline
      %\multicolumn{2}{*}{character-level} & He preened on a レ drink of copper. \\
      word-level & morphological & Higher education school {\bf xur} divided into six institutions.\\
      \hline
      \multirow{5}{*}{phrase-level} & \multirow{2}{*}{syntactic} & Swallowing hard, Verity stared {\bf at the these}, desperately wanting to see if they congealed.\\
      \cline{2-3}
      & \multirow{2}{*}{semantic} & The west section is a {\bf fig octagon}.\\
      \cline{3-3}
      & & A private apartment with nothing but \textbf{hot cooled} water.\\
      \cline{2-3}
      & \multirow{2}{*}{predication} &
       He already {\bf costumes his relationship} with my mother carefully.\\
      \cline{3-3}
       & & Voices rapped on the \textbf{incremental door}.\\
       \hline
       \multirow{4}{*}{sentence-level}
      & \multirow{1}{*}{out-of-context} & Like a {\bf cataract}, Horatius responds, ``You are better than me.''\\
      \cline{2-3}
      & \multirow{1}{*}{self-contradictory}  & The newspaper is published \textbf{weekly} and \textbf{biannually}.\\
      \cline{2-3}
      & \multirow{2}{*}{pragmatic} & She grew up with \textbf{three sisters} and \textbf{ten sisters}.\\
      \cline{3-3}
      & & It should apply between \textbf{the extreme} and \textbf{the extreme}.\\
      \hline
    \end{tabular}
    \caption{Examples of sentences sampled from BERT's prior that received low naturalness ratings from our participants, including sources forms of unnaturalness like predicability or category errors (e.g. doors typically do not have the property of ``incrementality''), semantic incoherence (``hot cooled water''), or contradictory constructions (especially for longer sentences). More examples can be found in table \ref{tab:more_examples} and in the online supplement.}
    \label{tab:examples}
\end{table*}

These deviations present a core theoretical and methodological puzzle for computational linguistics. 
How do we elicit and characterize the full \emph{prior}\footnote{We use the term \emph{prior} to refer to graded linguistic knowledge assigning probabilities to all possible sentences. While we focus on text, this prior is also the foundation for more grounded, pragmatic language use.} that a particular model has learned over possible sentences in a language?
A dominant approach has been to design benchmark suites that probe theoretically important aspects of the prior, and compare model behavior to human behavior on those tasks \cite[e.g.][]{warstadt2019blimp,ettinger2020bert}.
Yet this approach can be restrictive and piecemeal: it is not clear ahead of time which tasks will be most diagnostic, and many sources of ``weirdness'' are not easily operationalized \cite{kuribayashi2021lower}. 

A more holistic, bottom-up alternative is to directly examine samples from the model's prior and compare them against those from human priors. 
However, many successful models do not explicitly expose this distribution, and many generation methods optimize the ``best'' sentences rather than theoretically meaningful or representative ones.
For example, masked language models (MLMs) like BERT \cite{devlin2018bert} are \emph{dependency networks} \cite{heckerman2000dependency,toutanova2003feature}, trained to efficiently learn an independent collection of conditional distributions without enforcing consistency between them.
In other words, these conditionals may not correspond to any coherent joint distribution at all, leading recent work to focus on other score-based sampling objectives \cite{goyal2021exposing}.

Here, we explore the use of serial reproduction chains (see Fig.~\ref{fig:front}) to overcome these challenges. 
While a naive (pseudo-)Gibbs sampler is indeed problematic for MLMs, the literature on Generative Stochastic Networks \citep[GSNs;][]{bengio2014deep} has formally shown that a simple algorithmic variant we call \emph{GSN sampling} produces a stationary distribution that is, in fact, a unique and consistent estimator of the ground-truth joint distribution.
Furthermore, while the independent conditionals learned by dependency networks may be arbitrarily inconsistent in theory, empirical work has found that these deviations tend to be negligible in practice, especially on larger datasets \cite{heckerman2000dependency,neville2007relational}. 
Thus, we argue that it is both theoretically and empirically justified to take these samples as uniquely representative of the model's prior over language. 

%While \citet{wang2019bert} found that the sampled sentences are reasonably diverse and fluent, therefore validating Gibbs sampling as a practical method for generation, 
We begin in Section \ref{sec:approach} by introducing the serial reproduction approach and clarifying the problem of re-constructing a joint distribution from a dependency network.
We then validate that our chains are well-behaved (Section \ref{sec:validation}) and compare the statistics of samples from BERT's prior to the lexical and syntactic statistics of its ground-truth training corpus to measure distributional similarity (Section \ref{sec:distributional}).
Finally, in Section \ref{sec:judgments}, we present a large-scale behavioral study eliciting naturalness judgments from human speakers and identify features of the generated sentences which most strongly predict human ratings of ``weirdness'' (see Table \ref{tab:examples}).
We find that GSN samples closely approximate the ground-truth distribution and are judged to be more natural than other methods, while also revealing areas of improvement that have been difficult to quantify with top-down benchmarks. 

\section{Approach}
\label{sec:approach}

\subsection{Serial reproduction}

Our approach is inspired by serial reproduction games like Telephone, where an initial message is gradually relayed along a chain from one speaker to the next.
At each step, the message is changed subtly as a result of noisy transmission and reconstruction, and the final version of the message often differs drastically from the first. 
This serial reproduction method, initially introduced to psychology by \citet{bartlett1995remembering}, has become an invaluable tool for revealing human \emph{inductive biases} \cite{xu2010rational,langlois2021serial,sanborn2010uncovering,harrison2020gibbs}.
Because reconstructing a noisy message is guided by the listener's prior expectations, such chains eventually converge to a stationary distribution that is equivalent to the population's prior, reflecting what people expect others to say \cite{kalish2007iterated,griffiths2007language,beppu2009iterated}.
For example, \citet{meylan2021evaluating} recently evaluated the ability of neural language models to predict the changes made to sentences by human participants at each step of a serial reproduction chain.  
%The models' predictions gradually improved as the chains converged toward more representative language.
Thus, while serial reproduction is commonly used to probe \emph{human} priors, and to compare models against human data, it is not yet in wide use for probing the models themselves. 

\begin{figure}[t!]
\begin{center}
\includegraphics[width=0.9\linewidth]{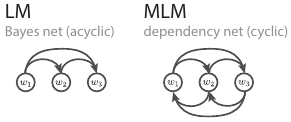}
\caption{While autoregressive language models (LMs) are Bayes nets, masked language models (MLMs) are dependency networks with cyclic dependencies.}
\label{fig:schematic}
\end{center}
\end{figure}

\subsection{BERT as a dependency network}

There has been considerable confusion in the recent literature over how to interpret the MLM objective used to train models like BERT, and how to interpret samples from such models. 
\citet{wang2019bert} initially observed that BERT was a Markov Random Field (MRF) and proposed a Gibbs sampler that iteratively masks and reconstructs different sites $k$ by sampling from the conditional given the tokens at all other sites $\hat{P}(w_k | w_{-k})$.
As observed by \citet{goyal2021exposing},
%\footnote{And corrected by the original authors in an earlier erratum: \href{https://kyunghyuncho.me/bert-has-a-mouth-and-must-speak-but-it-is-not-an-mrf/}{https://kyunghyuncho.me/bert-has-a-mouth-and-must-speak-but-it-is-not-an-mrf/}}
however, this procedure does not actually correspond to inference in the MRF. 
Unlike auto-regression language models (LMs) like GPT-3 \cite{brown2020language}, which define an acyclic dependency graph (or Bayes net) from left-to-right, MLMs have cyclic dependencies (see Fig.~\ref{fig:schematic}) and are therefore usefully interpreted as dependency networks rather than Bayes networks \cite{heckerman2000dependency}.
Because dependency networks estimate independent conditionals, there is no guarantee that these conditionals are consistent (i.e. they may violate Bayes rule) and therefore do not represent a coherent joint distribution. 

Still, it is possible to re-construct a joint distributions from these conditionals. 
For example, \citet{heckerman2000dependency} proved that if sites are visited in a fixed order, a (pseudo-)Gibbs chain similar to the one used by \citet{wang2019bert} does converge to a stationary distribution that is a well-formed joint. 
The problem is that different orders may yield different joint distributions, making it difficult to interpret any distributions as definitive. 
This ambiguity was resolved by the Generative Stochastic Network framework proposed by \citet{bengio2014deep}.
Instead of visiting sites in a fixed order, a GSN sampler randomly chooses which site to visit at each step (with replacement), thus preserving aperiodicity and ergodicity. 
Specifically, this algorithm begins by initializing with a sequence $\{w_1^{0}, \dots, w_n^{0}\}$.
At each step $t$ of the chain, we randomly choose a site $k\in{1,\dots,n}$ to mask out, and we sample a new value $w_k^{t+1}$ from the conditional distribution $P(w_k | w_{-k}^{t})$ with the other $n-1$ sites fixed.
%These steps are organized into \emph{epochs}, representing a single pass through all sites.
%We choose sites randomly without replacement, such that each epoch is exactly $n$ steps.

A key theorem of \citet{bengio2013generalized,bengio2014deep} proves that the stationary distribution arising from the GSN sampler defines a unique joint distribution, and furthermore, this stationary distribution is a consistent estimator of the ground-truth joint distribution\footnote{Technically, the proof only holds if the dependency network was trained using consistent estimators for the conditionals, which is the case for the cross-entropy loss used by BERT; see also \citet{bertconsistency}.}.
%They learn to represent conditional distributions of a masked token $w_k$ in a sequence, conditioned on all other tokens: $P(w_k | w_{-k})$.
%This set of conditional distributions is precisely what is needed to define a Gibbs sampler  for the joint distribution $P(w_1, \dots, w_n)$, which represents the model's prior over sentences.
 %\cite[see][for an introduction to MCMC methods]{andrieu2003introduction,murphy2012machine}. 
Importantly, this stationary distribution differs from the one given by the Metropolis-Hastings (MH) approach suggested by \citet{goyal2021exposing}, which uses the GSN sampler as a \emph{proposal} distribution but accepts or rejects proposals based on an energy-based pseudo-likelihood defined by the sum of the conditional scores at each location \cite{salazar2019masked}. 
This MH sampler instead converges to an implicit stationary distribution defined by the energy objective\footnote{Although our focus is on evaluation rather than algorithmic performance characteristics, we note that because GSN sampling does not require calculating energy scores to determine the acceptance probability for each sample, it is significantly faster, especially for longer sequences.}.

\subsection{Mixture kernels}

In practice, Markov chain sampling methods have many failure modes.
Most prominently, because samples in the chains are not independent, it is challenging to guarantee convergence to a stationary distribution, and the chain is easily ``stuck'' in local regions of the sample space \cite{gelman1992inference}.
Typically, samples from a \emph{burn-in} period (e.g. the first $m$ epochs) are discarded to reduce dependence on the initial state, and a \emph{lag} between samples (e.g. recording only every $l$ epochs) is introduced to reduce auto-correlation.
However, the problem is particularly severe for language models like BERT where there are strong mutual dependencies between words at different sites. 
For example, once the chain reaches a tri-gram like `Papua New Guinea', it is unlikely to change any single word while keeping the other words constant.
To ensure ergodicity, we use a mixture kernel introducing a small constant probability ($\epsilon = 0.001$) of returning to the initial distribution of \texttt{[MASK]} tokens on each epoch, allowing the chain to burn in again.

\section{Validating the stationary distribution}
\label{sec:validation}

In this section, we validate that the samples produced by our serial reproduction method are representative of the stationary prior distribution. 
More specifically, we consider two basic properties of the chain: \emph{convergence} and \emph{independence}.
For these analyses, we consider samples from the pretrained \texttt{bert-base-uncased} model with 12 layers, 12 heads, and 110M parameters\footnote{\href{https://huggingface.co/bert-base-uncased}{https://huggingface.co/bert-base-uncased}}.

\subsection{Convergence}

We begin by checking the convergence time for chains generated by GSN sampling.
Theoretical bounds derived for serial reproduction chains give a convergence time of $n \log n$, where $n$ is the number of sites \cite[see][]{rafferty2014analyzing}.
To check these convergence bounds in practice, we set $n=21$ and select 20 sentences from Wikipedia to serve as initial states, and run 10 chains initialized at each sentence.
We ensured that half of these sentences have high initial probability (under BERT's energy score) and half have low initial probability.
We find that these distributions indeed begin to quickly mix in probability (see Figure \ref{fig:squeeze}).
Because longer sentences may require a longer burn-in time, we conservatively set our burn-in window to $m=1000$ epochs for our subsequent experiments.

\subsection{Independence}

Second, we want to roughly ensure independence of samples, so that the statistics of our distribution of samples isn't simply reflecting auto-correlation in the chain.
For a worst-case analysis of a local minimum, suppose $P(w_i | w_{-i}) < \delta$ $(0<\delta<1)$ for all $i \in [1, \dots, k]$, where $k$ is the sentence length in tokens.
Then the probability of re-sampling the same sentence is roughly $<\delta^{k \cdot n}$ after $n$ epochs.
We can solve for the number of epochs $n$ we need to bound the probability of re-sampling the exact same sentence under $\epsilon$ for a given worst-case $\delta$.
For example, if $\delta = 0.99$ and we want to ensure that the probability of re-sampling the same sentence is below a threshold $\epsilon=0.01$, then $n=47$ epochs will likely suffice. 
Ensuring complete turnover in the worst case scenario requires much longer lags, i.e. $[1-(1-\delta)^{k}]^n < \epsilon$.

To evaluate the extent to which these cases arise in practice, we examine auto-correlation rates  on longer chains (50,000 epochs). 
We calculate correlations between the energy scores at each epoch as a proxy for the state: when the chain gets stuck re-sampling the same sentence, the same scores appear repeatedly.
We find that auto-correlation is generally high, but our mixture kernel prevents the worst local minima for both the MH chain \cite{goyal2021exposing} and our GSN chain (see Fig.~\ref{fig:ac}), although we still found higher auto-correlation rates for the MH chain. 
To further examine these minima, we examined edit rates: the number of changes made to the sentence within an epoch.
Without the mixture kernel, we observe long regions of consistently low edit rates (e.g. in some cases, 5000 epochs in a row of exactly the same sentence) which disappear under the mixture kernel (see Fig.~\ref{fig:editrates}). 

\begin{figure*}[thb!]
\begin{center}
\includegraphics[width=0.48\linewidth]{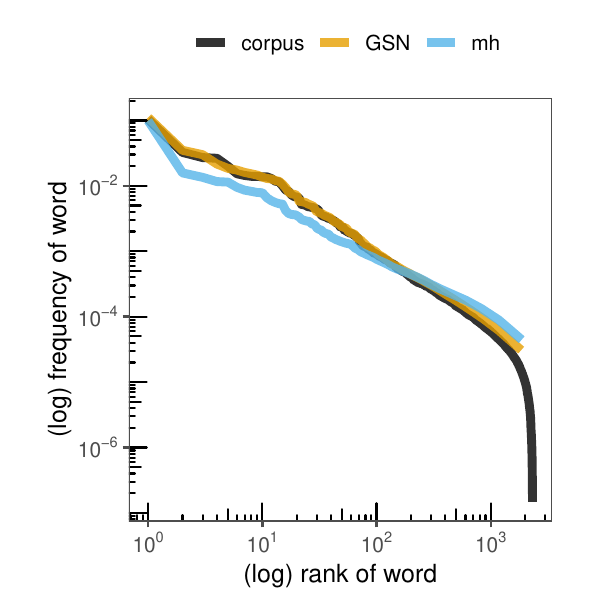}
\includegraphics[width=0.48\linewidth]{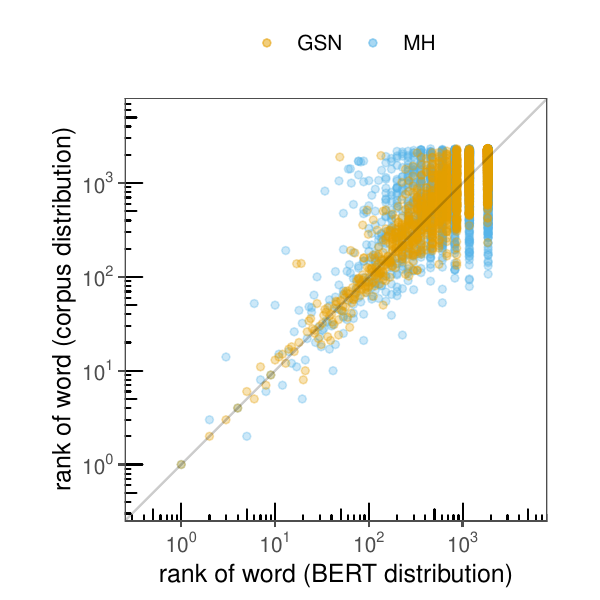}
\caption{The lexical frequencies  of our GSN samples (A) closely match the Zipfian distribution of the corpus and (B) closely correlate with the corresponding frequencies of the corpus distribution.}
\label{fig:lex}
\end{center}
\end{figure*}
Based on these observations, we set the lag to $l=500$ epochs to maintain relatively high independence between samples.

% Third, we want to check what proportion of generated BERT sentences are just memorized from its training corpus. 
% The intractable/brute-force way to do this is a sort of BLEU score, i.e. for each sentence in BERT corpus, we search over entire Wikipedia corpus and measure how many tokens overlap (ensuring same vocabulary). 
% Then if any BERT sentence has an exact match or a close-to-exact match, it's a candidate for memorization.
% The slightly more efficient way to do this is to take nearest-neighbors in an embedding space?
% Maybe there's another way?
% \red{Fuzzy tri-gram multiset match, e.g. Carlini et al.? https://arxiv.org/pdf/2012.07805.pdf}

\section{Distributional comparisons}
\label{sec:distributional}

In this section, we examine the extent to which higher-order statistics of sentences from BERT's prior are well-calibrated to the data it was trained on.
This kind of comparison provides a richer sense of what the model has learned or failed to learn than traditional scalar metrics like perplexity \cite{takahashi2017neural,meister2021language,takahashi2019evaluating,pillutla2021mauve}.

\subsection{Corpus preparation}
\begin{figure*}[t]
\begin{center}
\includegraphics[width=0.48\linewidth]{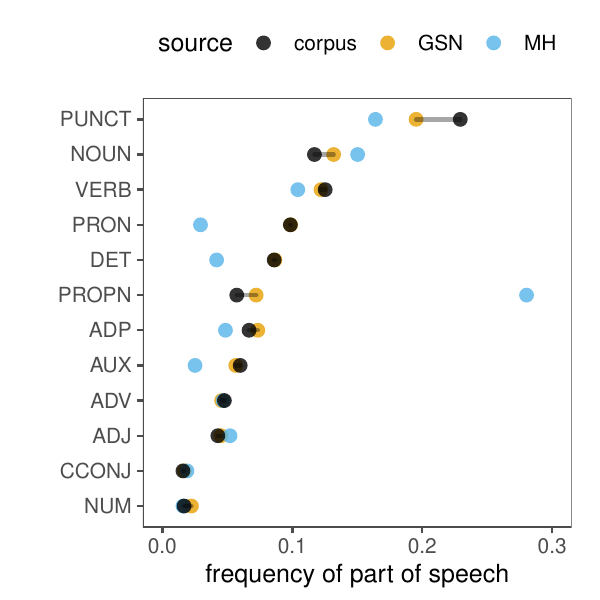}
\includegraphics[width=0.48\linewidth]{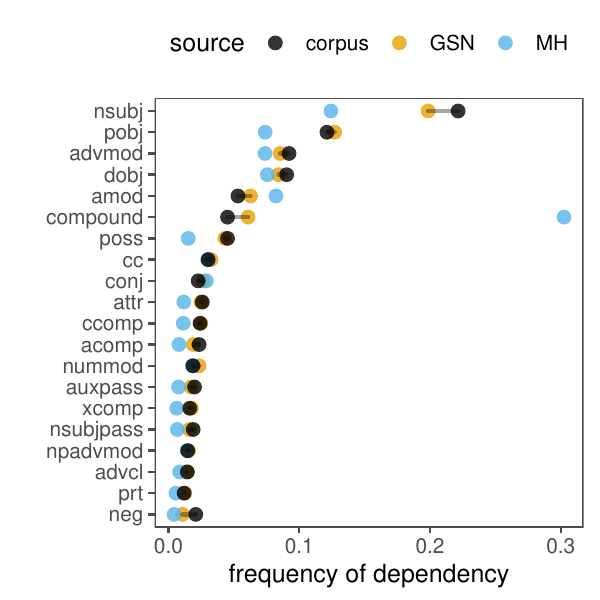}
\caption{The relative frequencies of different parts of speech (left) and dependencies (right) in the ground-truth training corpora closely matched for GSN samples. In all cases, the GSN frequencies fell closer to the ground-truth than the MH frequencies.}
\label{fig:pos}
\end{center}
\end{figure*}
The version of BERT we analyzed in the previous section was trained on a combination of two corpora: \emph{Wikipedia} and \emph{BookCorpus}.
In order to make valid comparisons between human priors and machine priors, we needed to closely match BERT-generated sentences with a comparable subset of human-generated sentences from these combined corpora.
There are two technical challenges we must overcome to ensure comparable samples, concerning the \emph{sentencizer} and \emph{tokenizer} steps.

First, because our unit of comparison is the \emph{sentence}, we needed to control for any artifacts that may be induced by how we determine what sentences are (e.g. if our Wikipedia sentences were systematically split on abbreviations, skewing the distribution toward fragments). 
We therefore applied the same \texttt{punkt} sentencizer to create our distribution of Wikipedia sentences and to check our BERT samples for cases where the generated sequence contained multiple sentences or ended with a colon or semicolon.

Second, we needed a tokenizer that equates sentence length.
Because bi-directional models like BERT operate over sequences of fixed length, all samples drawn from a single chain have the same number of tokens.

Critically, however, BERT chains are defined over sequences of \emph{WordPiece} tokens, so once these sequences are decoded back into natural language text, they may yield sentences of varying length, depending on how the sub-word elements are combined together\footnote{One additional complexity is that the mapping between WordPiece tokens and word tokens is non-injective. There exist multiple sequences of sub-word tokens that render to the same word (e.g. the WordPiece vocabulary contains a token for the full word `missing' but it is also able to generate `missing' by combining the sub-word tokens `miss'+`\#ing'). However, these cases are rare.} (see Fig.~\ref{fig:misalignment}).
We solve this alignment problem by using the \emph{WordPiece} tokenizer to extract sentences of fixed sub-word token length from our text corpora, yielding equivalence classes of corpus sentences that are all tokenized to the same number of WordPiece tokens.
We ran GSN and MH chains over sentences of $n=11$ tokens, representing the modal lengths of sentences in BookCorpus (see Fig.~\ref{fig:wikisentlength}). 
We obtained 5,000 independent sentences from each sampling method after applying our conservative burn-in and lag, and combined the Wikipedia and BookCorpus sentences together into a single corpus that is representative of BERT's training regime.

\subsection{Lexical distributions}

We begin by comparing the \emph{lexical frequency} statistics of our samples from BERT against the ground-truth corpus statistics.
First, we note that the relationship between rank and frequency of tokens in the GSN sampling matches the Zipfian distribution of its training corpus better than those produced by MH sampling (see Fig.~\ref{fig:lex}A).
However, it is possible to produce the same overall distribution without matching the empirical frequencies of individual words.
We next examined the respective ranks of each word across the two distributions. 
Overall, the word ranks in the GSN samples had a strong Spearman rank correlation of $r=0.75$ with the word ranks in the ground-truth corpus; the MH samples had a significantly lower correlation of $r=0.48$ (Pearson $z = 17, p< 0.001$, Fig ~\ref{fig:lex}B).
Most disagreements lay in the tails where frequency estimates are particularly poor (e.g. many words only appeared once in our collection of samples). 
Indeed, among words with greater than 10 occurrences, the correlation improved to $r=0.83$ for GSN and $r=0.65$ for MH.

To understand this relationship further, we conducted an error analysis of lexical items which were systematically over- or under-produced by BERT relative to its training corpus. 
We found that certain punctuation tokens (e.g. parentheses) were over-represented in both the GSN samples and the MH samples, while contractions like \emph{’s} and \emph{’d} were under-represented.
%as well as temporal words (\emph{now}, \emph{around}, \emph{during}, \emph{years}, \emph{before}, \emph{time}) and comparatives (e.g. \emph{than}, \emph{more}). 
The MH samples specifically over-produced proper names such as \emph{Nina} and \emph{Jones}.
Finally, due to the use of sub-word representations, we found a long tail of morphologically complex words that did not appear at all in the training corpus (e.g. names like Kyftenberg or Streckenstein and seemingly invented scientific terms like lymphoplasmic, neopomphorus, or pyranolamines). 

\subsection{Syntactic distributions}

While the lexical distributions were overall well-matched for GSN samples, our error analysis suggested potential structure in the deviations. 
In other words, entire grammatical \emph{constructions} may be over- or under-represented, not just particular words. 
To investigate these patterns, we used the \texttt{spacy} library to extract the parts of speech and dependency relations that are present within each sentence. 
We are then able to examine, in aggregate, whether certain classes of constructions are disproportionately responsible for deviations. 
Our findings are shown in Fig.~\ref{fig:pos}. 
Overall, the distributions are close, but several areas of misalignment emerge. 
For parts of speech, we observe that the GSN sampler is slightly over-producing nouns (and proper nouns) while under-producing verbs and prepositions.
We also observe that it is over-producing noun-related dependencies (e.g. compound nouns and appositional modifiers, which are noun phrases modifying other noun phrases, as in ``Bill, my brother, visited town'').
This pattern suggests that BERT's prior may be skewed toward (simpler) noun phrases while neglecting more complex constructions. 

\subsection{Sentence complexity}

\begin{figure}[t]
\begin{center}
\vspace{-2em}
\includegraphics[width=1\linewidth]{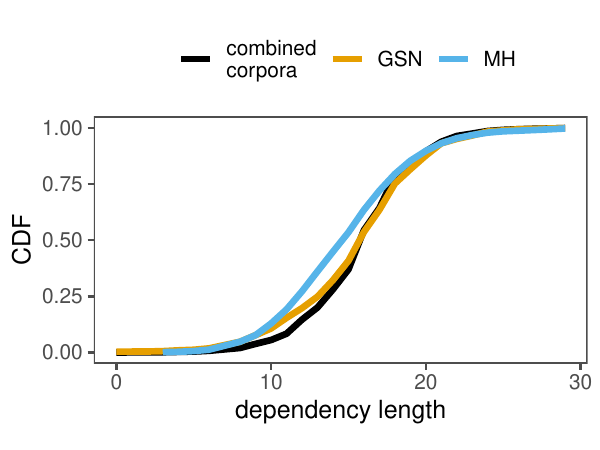}
\caption{Cumulative probability distribution of dependency lengths across sentences from BERT chains and from the training corpus.}
\vspace{-1em}
\label{fig:dep_dist_cdf}
\end{center}
\end{figure}

One hypothesis raised by comparing distributions of syntactic features is that BERT may be \emph{regularizing} the complex structure of its input toward simpler constructions. 
To test this hypothesis, we operationalize syntactic complexity using a measure known as the average \emph{dependency length} of a sentence \cite{futrell2015large,grodner2005consequences}. 
This measure captures the (linear) distance between syntactically related words, which increases with more complex embedded phrase structures. 
We found that the distribution of dependency distances in the sentences produced by GSN sampling is overall more similar to those in its training corpus than the MH (Fig.~\ref{fig:dep_dist_cdf}), although closer analysis suggests it is still skewed slightly simpler (see Fig.~\ref{fig:dependency_distance}). 

\section{Human judgments}
\label{sec:judgments}
Finally, while our corpus comparisons highlighted particular ways in which samples from BERT's prior were well-calibrated to the high-level statistics of its training distribution, it is unclear whether these agreements or deviations `matter' in terms of naturalness. 
In this section, we elicit human naturalness judgments in order to provide a more holistic measure of potential `weirdness' with BERT sentences.

\subsection{Experimental methods}

We recruited 1016 fluent English speakers on the Prolific platform and asked them to judge the naturalness of 4040 unique sentences from three length classes: short (11 tokens), medium (21 tokens), and long (37 tokens).
1675 of these sentences were from the stationary state of the different chains, 2339 were from the burn-in phase (i.e. $<1000$ epochs), and the remainder were baseline sentences (149 from Wikipedia, 48 from a 5-gram model, and 42 from an LSTM model; see Appendix for details).
Each participant was shown a sequence of 25 sentences in randomized order, balanced across different properties of the stimulus set\footnote{In a later batch, we increased the number of sentences per participant to 40. The task was approximately 10 minutes and participants were paid \$2.50, for an average compensation rate of \$15/hr.}. 
On each trial, one of these sentences appeared with a slider ranging from 0 (``very weird'') to 100 (``completely natural'')\footnote{See \citet{clark2021all} for a discussion of the merits of phrasing the question in terms of naturalness instead of asking participants to judge whether it was produced by a human or machine.}.
After excluding 8 participants who failed the attention check (i.e. failed to rate a scrambled sentence below the midpoint of the scale and a human-generated sentence above the midpoint), we were left with an average of 7.3 responses per sentence. 

\subsection{Behavioral results}

\begin{figure}[t!]
\begin{center}
\includegraphics[width=0.99\linewidth]{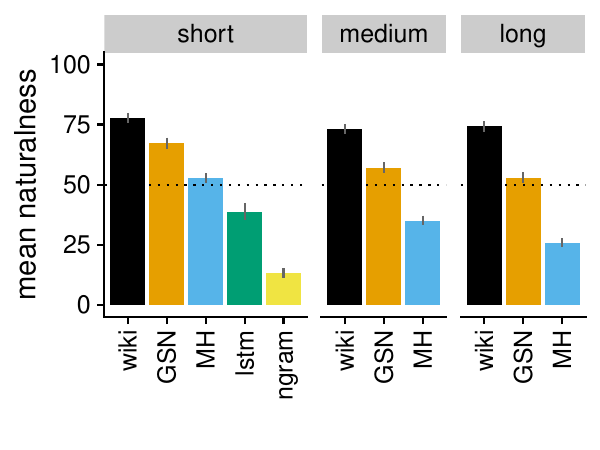}
\vspace{-3em}
\caption{Empirical naturalness ratings elicited from the stationary GSN distribution, compared to different baselines at different sentence lengths. Error bars are bootstrapped 95\% CIs.}
\vspace{-1em}
\label{fig:empirical}
\end{center}
\end{figure}

We begin by comparing the naturalness of sentences from the stationary GSN distribution to other baselines (see Fig.~\ref{fig:empirical}), using a linear regression model predicting trial-by-trial judgments as a function of categorical variables encoding sentence length (short, medium, long) and the source of the sentence (Wikipedia, GSN, MH, LSTM, or n-gram).
First, we find that the naturalness of sentences from GSN declines by 14 points at longer sentence lengths, $p <0.001$, while the naturalness of Wikipedia sentences is unaffected by length (interaction term, $p< 0.001$), consistent with results reported by \citet{ippolito2020automatic}.
Furthermore, among short sentences, where we included additional baselines, we find that GSN sentences tend to be rated as slightly less natural than sentences from Wikipedia (+10 points, $p<0.001)$ but more natural than those produced by an n-gram model (-52 points, $p < 0.001$), LSTM model (-25 points, $p <0.001$); or MH sampling from the same BERT conditionals (-15 points, $p<0.001$; see Table \ref{tab:behaviorregression}). 
MH samples also deteriorate significantly in naturalness for longer sentences compared to GSN samples ($p < 0.001$). 
Finally, we examine naturalness ratings across the the burn-in period, finding that ratings decline steadily across the board as the chain takes additional steps (linear term: $t(7297)=-12.4, p <0.001$), suggesting gradual deviation away from the initial distribution of Wikipedia sentences toward the stationary distribution (shown as the green and grey regions, respectively, in Fig.~\ref{fig:burnin}).
% Next, we examined the dynamics of the burn-in period of the chain (epoch $<1000$) using another mixed-effects model including fixed effects of the (log) burn-in step (both linear and quadratic), sentence length, (short, medium, long) and the probability class of the initial state (high vs. low; see Figure \ref{fig:squeeze}). 
% However, this relationship is complicated by the initial state. 
% While chains initialized in high probability states decline in (log) linear time, chains initialized in low probability states have a shallower, more quadratic relationship, getting less natural before ultimately converging to the same stationary distribution (linear interaction term: $t(7144) = -5.0, p < 0.001$; quadratic interaction term: $t(7276) = -5, p < 0.001$). 
% This result supports our analysis of burn-in in Section $\ref{sec:validation}$, and suggests that the model may be less well-calibrated for lower-probability regions. 

\subsection{Predicting naturalness}

Given that sentences from the stationary GSN distribution are judged to be less natural than human-generated sentences overall, we are interested in explaining \emph{why}. 
Which properties of these sentences make them sound strange? 
We approach this problem by training a regression model to predict human judgments from attributes of each sentence. 
We include all part of speech tag counts and dependency counts, as well as the sentence probability scored under BERT, and the sentence length.
We use a cross-validated backwards feature selection procedure to select the most predictive set of these features for a linear regression \cite{kuhn2013introduction}\footnote{Specifically,  we used the \texttt{lmStepAIC} procedure implemented in the \texttt{caret} R package, with $k=10$ folds.}.

The best-fitting model used 26 features and achieved an (adjusted) $R^2 = 0.21$.
The only features associated with significantly \emph{lower} ratings were the use of adpositions (e.g. \emph{before}, \emph{after}) and coordinating conjunctions.
%Features associated with significantly \emph{higher} ratings included higher sentence probability and dependency distances, as well as 12 different dependencies (case-marking elements, the number of conjuncts, noun compounds, interjections, passive and active subjects, relative clauses and open clausal complements, parataxis, as well as adjectival,  numerical, and quantifier phrase modifiers).
Importantly, we found that including a categorical variable of corpus (i.e. Wikipedia vs. GSN) significantly improved model fit even after controlling for all other features, $\chi^2(1)=7135, p < 0.001$, suggesting that sources of ``weirdness'' are not being captured by typical statistics.
We show some of these low-naturalness sentences in Table \ref{tab:examples} and \ref{tab:more_examples}.

\section{Discussion}

\subsection{Probing through generation}

A core idea of our serial reproduction approach is to use generation as a window into a model's prior over language. 
While a variety of metrics and techniques have been proposed to quantify the ``quality'' of generation, especially in the domains of open-ended text generation and dialogue systems \cite{caccia2018language,li2019don,guidotti2018survey,celikyilmaz2020evaluation}, these metrics have typically been applied to compare specific generation algorithms and operationalize specific pitfalls, such as incoherence, excess repetition, or lack of diversity.
Consequently, it has been difficult to disentangle the extent to which deviations resulting from generations are an artifact of specific decoding algorithms (e.g. greedy search vs. beam search) or run deeper, into the prior itself.
For the purposes of probing, we suggest that it is important to ask not only how to generate the highest-scoring sentences but how to generate sentences that may be interpreted as representative of the model's prior, as formal results on GSNs have effectively provided.

\subsection{GSN vs. energy-based objectives}

We found that the prior distribution yielded by the GSN sampler more closely approximated the lexical and syntactic distributions of the ground-truth corpus and also sounded more ``natural'' to humans than the samples yielded by MH.
These results are in contrast to findings by \citet{goyal2021exposing}, showing that MH produced high-quality BLEU scores on a Machine Translation (MT) task compared to a degenerate (pseudo-)Gibbs sampler.
There are several possible reasons for this discrepancy. 
One possibility may be task-specific: while we focused on unconditional generation, \citet{goyal2021exposing} focused on a neural machine translation (MT) task, where sentence generation was always conditioned on a high-quality source text and thus remained within a constrained region of sentence space. 
Another possibility is that we ran substantially longer chains (50,000 epochs compared to only 33 epochs) and the pitfalls of MH sampling only emerged later in the chain.

More broadly, our corpus comparisons and human evaluations suggest serious limitations of simple ``quality'' metrics like energy values. 
We found that the best-scoring states were often degenerate local minima with mutually supporting n-grams (such as repetitive phases and names like ``Papua New Guinea'').
Indeed, there was only a loose relationship between energy scores and participants' judgments in our study, with many poorer-scoring sentences judged to be more natural than better-scoring sentences (e.g. overall, the distribution of Wikipedia sentences tended to be much lower-scoring under the energy function despite being rated as more natural).
We empirically validated that the stationary distribution of the GSN chain successfully approximates even higher-order statistics of the ground-truth corpus, suggesting that the raw conditionals of the dependency network may implicitly acquire the joint distribution, without requiring guarantees of consistency. 
%This free generation scheme may have revealed parts of the sentence space where having better BERT scores does not mean sounding more ``natural''.

\subsection{Other architectures}

Serial reproduction methods are particularly useful for probing models that do not directly generate samples from their prior.
For auto-regressive models like GPT-2, these samples are obtained more directly by running the model forward \cite[and, indeed, ancestral sampling produces text that better balances the precision-recall tradeoff than other algorithms;][]{pillutla2021mauve}.
While we focused on BERT, this method may be particularly useful for encoder-decoder architectures
like BART \cite{lewis2019bart} which more closely resemble the human Telephone Game task, requiring full reconstruction of the entire sentence from noisy input rather than reconstruction of a single missing word. 
Indeed, these architectures may overcome an important limitation of serial reproduction with BERT: because these chains operate over a fixed sequence length, the resulting prior is not over all of language but only over sentences with the given number of WordPiece tokens.
Finally, while we focused on unconditional generation, the GSN sampler also generalizes straightforwardly to conditional generation, where a subset of sites are fixed and the masked site is chosen from the remaining set.

\subsection{Conclusions}

Serial reproduction paradigms have been central for exposing human priors in the cognitive sciences.
In this paper, we drew upon the theory of iterated learning and of Generative Stochastic Networks (GSNs) to expose the priors of large neural language models, which are often similarly inscrutable.
We hope future work will consider other points of contact between these areas and draw more extensively from the theory developed to understand dependency networks.
%For example, more extensive forms of reproduction may be studied, where the output produced by one model is used to \emph{train} another model, gradually revealing inductive biases implicit in the architecture and training regime.
More broadly, as language models become increasingly adaptive and deployed in increasingly unconstrained settings, bottom-up probing has the potential to reveal a broader spectrum of ``weirdness'' than top-down evaluative benchmarks.

\section*{Acknowledgements}

This work was supported by NSF grant \#1911835 to RDH. We are grateful to Jay McClelland, Adele Goldberg, and Stephan Meylan for helpful conversations, and to three anonymous reviewers for feedback that improved our work.  

\bibliography{emnlp2020}

\begin{thebibliography}{53}
\expandafter\ifx\csname natexlab\endcsname\relax\def\natexlab#1{#1}\fi

\bibitem[{Bartlett(1932)}]{bartlett1995remembering}
Frederic~Charles Bartlett. 1932.
\newblock \emph{Remembering: A study in experimental and social psychology}.
\newblock Cambridge University Press.

\bibitem[{Bender et~al.(2021)Bender, Gebru, McMillan-Major, and
  Shmitchell}]{bender2021dangers}
Emily~M Bender, Timnit Gebru, Angelina McMillan-Major, and Shmargaret
  Shmitchell. 2021.
\newblock On the dangers of stochastic parrots: Can language models be too big?
\newblock In \emph{Proceedings of the 2021 ACM Conference on Fairness,
  Accountability, and Transparency}, pages 610--623.

\bibitem[{Bengio et~al.(2014)Bengio, Laufer, Alain, and
  Yosinski}]{bengio2014deep}
Yoshua Bengio, Eric Laufer, Guillaume Alain, and Jason Yosinski. 2014.
\newblock Deep generative stochastic networks trainable by backprop.
\newblock In \emph{International Conference on Machine Learning}, pages
  226--234. PMLR.

\bibitem[{Bengio et~al.(2013)Bengio, Yao, Alain, and
  Vincent}]{bengio2013generalized}
Yoshua Bengio, Li~Yao, Guillaume Alain, and Pascal Vincent. 2013.
\newblock Generalized denoising auto-encoders as generative models.
\newblock \emph{Advances in neural information processing systems}, 26.

\bibitem[{Beppu and Griffiths(2009)}]{beppu2009iterated}
Aaron Beppu and Thomas Griffiths. 2009.
\newblock Iterated learning and the cultural ratchet.
\newblock In \emph{Proceedings of the Annual Meeting of the Cognitive Science
  Society}, volume~31.

\bibitem[{Braverman et~al.(2020)Braverman, Chen, Kakade, Narasimhan, Zhang, and
  Zhang}]{braverman2020calibration}
Mark Braverman, Xinyi Chen, Sham Kakade, Karthik Narasimhan, Cyril Zhang, and
  Yi~Zhang. 2020.
\newblock Calibration, entropy rates, and memory in language models.
\newblock In \emph{International Conference on Machine Learning}, pages
  1089--1099. PMLR.

\bibitem[{Brown et~al.(2020)Brown, Mann, Ryder, Subbiah, Kaplan, Dhariwal,
  Neelakantan, Shyam, Sastry, Askell et~al.}]{brown2020language}
Tom~B Brown, Benjamin Mann, Nick Ryder, Melanie Subbiah, Jared Kaplan, Prafulla
  Dhariwal, Arvind Neelakantan, Pranav Shyam, Girish Sastry, Amanda Askell,
  et~al. 2020.
\newblock Language models are few-shot learners.
\newblock \emph{Advances in Neural Information Processing Systems, 34}.

\bibitem[{Caccia et~al.(2020)Caccia, Caccia, Fedus, Larochelle, Pineau, and
  Charlin}]{caccia2018language}
Massimo Caccia, Lucas Caccia, William Fedus, Hugo Larochelle, Joelle Pineau,
  and Laurent Charlin. 2020.
\newblock Language {GAN}s falling short.
\newblock \emph{International Conference on Learning Representations}.

\bibitem[{Caliskan et~al.(2017)Caliskan, Bryson, and
  Narayanan}]{caliskan2017semantics}
Aylin Caliskan, Joanna~J Bryson, and Arvind Narayanan. 2017.
\newblock Semantics derived automatically from language corpora contain
  human-like biases.
\newblock \emph{Science}, 356(6334):183--186.

\bibitem[{Celikyilmaz et~al.(2020)Celikyilmaz, Clark, and
  Gao}]{celikyilmaz2020evaluation}
Asli Celikyilmaz, Elizabeth Clark, and Jianfeng Gao. 2020.
\newblock Evaluation of text generation: A survey.
\newblock \emph{arXiv preprint arXiv:2006.14799}.

\bibitem[{Clark et~al.(2021)Clark, August, Serrano, Haduong, Gururangan, and
  Smith}]{clark2021all}
Elizabeth Clark, Tal August, Sofia Serrano, Nikita Haduong, Suchin Gururangan,
  and Noah~A. Smith. 2021.
\newblock \href {https://doi.org/10.18653/v1/2021.acl-long.565} {All that{'}s
  {`}human{'} is not gold: Evaluating human evaluation of generated text}.
\newblock In \emph{Proceedings of the 59th Annual Meeting of the Association
  for Computational Linguistics.}, pages 7282--7296.

\bibitem[{Dasgupta et~al.(2020)Dasgupta, Guo, Gershman, and
  Goodman}]{dasgupta2020analyzing}
Ishita Dasgupta, Demi Guo, Samuel~J Gershman, and Noah~D Goodman. 2020.
\newblock Analyzing machine-learned representations: A natural language case
  study.
\newblock \emph{Cognitive Science}, 44(12):e12925.

\bibitem[{Devlin et~al.(2018)Devlin, Chang, Lee, and
  Toutanova}]{devlin2018bert}
Jacob Devlin, Ming-Wei Chang, Kenton Lee, and Kristina Toutanova. 2018.
\newblock {BERT}: Pre-training of deep bidirectional transformers for language
  understanding.
\newblock In \emph{Proceedings of NAACL-HLT}.

\bibitem[{Ettinger(2020)}]{ettinger2020bert}
Allyson Ettinger. 2020.
\newblock What bert is not: Lessons from a new suite of psycholinguistic
  diagnostics for language models.
\newblock \emph{Transactions of the Association for Computational Linguistics},
  8:34--48.

\bibitem[{Futrell et~al.(2015)Futrell, Mahowald, and Gibson}]{futrell2015large}
Richard Futrell, Kyle Mahowald, and Edward Gibson. 2015.
\newblock Large-scale evidence of dependency length minimization in 37
  languages.
\newblock \emph{Proceedings of the National Academy of Sciences},
  112(33):10336--10341.

\bibitem[{Garg et~al.(2018)Garg, Schiebinger, Jurafsky, and Zou}]{garg2018word}
Nikhil Garg, Londa Schiebinger, Dan Jurafsky, and James Zou. 2018.
\newblock Word embeddings quantify 100 years of gender and ethnic stereotypes.
\newblock \emph{Proceedings of the National Academy of Sciences},
  115(16):E3635--E3644.

\bibitem[{Gelman et~al.(1992)Gelman, Rubin et~al.}]{gelman1992inference}
Andrew Gelman, Donald~B Rubin, et~al. 1992.
\newblock Inference from iterative simulation using multiple sequences.
\newblock \emph{Statistical science}, 7(4):457--472.

\bibitem[{Goyal et~al.(2021)Goyal, Dyer, and
  Berg-Kirkpatrick}]{goyal2021exposing}
Kartik Goyal, Chris Dyer, and Taylor Berg-Kirkpatrick. 2021.
\newblock Exposing the implicit energy networks behind masked language models
  via {Metropolis--Hastings}.
\newblock \emph{arXiv preprint arXiv:2106.02736}.

\bibitem[{Griffiths and Kalish(2007)}]{griffiths2007language}
Thomas~L Griffiths and Michael~L Kalish. 2007.
\newblock Language evolution by iterated learning with {Bayesian} agents.
\newblock \emph{Cognitive Science}, 31(3):441--480.

\bibitem[{Grodner and Gibson(2005)}]{grodner2005consequences}
Daniel Grodner and Edward Gibson. 2005.
\newblock Consequences of the serial nature of linguistic input for sentenial
  complexity.
\newblock \emph{Cognitive Science}, 29(2):261--290.

\bibitem[{Guidotti et~al.(2018)Guidotti, Monreale, Ruggieri, Turini, Giannotti,
  and Pedreschi}]{guidotti2018survey}
Riccardo Guidotti, Anna Monreale, Salvatore Ruggieri, Franco Turini, Fosca
  Giannotti, and Dino Pedreschi. 2018.
\newblock A survey of methods for explaining black box models.
\newblock \emph{ACM computing surveys (CSUR)}, 51(5):1--42.

\bibitem[{Gulordava et~al.(2018)Gulordava, Bojanowski, Grave, Linzen, and
  Baroni}]{gulordava2018colorless}
Kristina Gulordava, Piotr Bojanowski, Edouard Grave, Tal Linzen, and Marco
  Baroni. 2018.
\newblock Colorless green recurrent networks dream hierarchically.
\newblock In \emph{Proceedings of NAACL-HLT}, page 1195–1205.

\bibitem[{Harrison et~al.(2020)Harrison, Marjieh, Adolfi, van Rijn,
  Anglada-Tort, Tchernichovski, Larrouy-Maestri, and
  Jacoby}]{harrison2020gibbs}
Peter Harrison, Raja Marjieh, Federico Adolfi, Pol van Rijn, Manuel
  Anglada-Tort, Ofer Tchernichovski, Pauline Larrouy-Maestri, and Nori Jacoby.
  2020.
\newblock Gibbs sampling with people.
\newblock \emph{Advances in Neural Information Processing Systems}, 33.

\bibitem[{Hawkins et~al.(2020)Hawkins, Yamakoshi, Griffiths, and
  Goldberg}]{hawkins2020investigating}
Robert Hawkins, Takateru Yamakoshi, Thomas Griffiths, and Adele Goldberg. 2020.
\newblock \href {https://doi.org/10.18653/v1/2020.emnlp-main.376}
  {Investigating representations of verb bias in neural language models}.
\newblock In \emph{Proceedings of the 2020 Conference on Empirical Methods in
  Natural Language Processing (EMNLP)}, pages 4653--4663.

\bibitem[{Heafield(2011)}]{heafield2011kenlm}
Kenneth Heafield. 2011.
\newblock Kenlm: Faster and smaller language model queries.
\newblock In \emph{Proceedings of the sixth workshop on statistical machine
  translation}, pages 187--197.

\bibitem[{Heckerman et~al.(2000)Heckerman, Chickering, Meek, Rounthwaite, and
  Kadie}]{heckerman2000dependency}
David Heckerman, David~Maxwell Chickering, Christopher Meek, Robert
  Rounthwaite, and Carl Kadie. 2000.
\newblock Dependency networks for inference, collaborative filtering, and data
  visualization.
\newblock \emph{Journal of Machine Learning Research}, 1(Oct):49--75.

\bibitem[{Holtzman et~al.(2019)Holtzman, Buys, Du, Forbes, and
  Choi}]{holtzman2019curious}
Ari Holtzman, Jan Buys, Li~Du, Maxwell Forbes, and Yejin Choi. 2019.
\newblock The curious case of neural text degeneration.
\newblock \emph{International Conference on Learning Representations}.

\bibitem[{Ippolito et~al.(2020)Ippolito, Duckworth, Callison-Burch, and
  Eck}]{ippolito2020automatic}
Daphne Ippolito, Daniel Duckworth, Chris Callison-Burch, and Douglas Eck. 2020.
\newblock Automatic detection of generated text is easiest when humans are
  fooled.
\newblock In \emph{Proceedings of the 58th Annual Meeting of the Association
  for Computational Linguistics}, pages 1808--1822.

\bibitem[{Kalish et~al.(2007)Kalish, Griffiths, and
  Lewandowsky}]{kalish2007iterated}
Michael~L Kalish, Thomas~L Griffiths, and Stephan Lewandowsky. 2007.
\newblock Iterated learning: Intergenerational knowledge transmission reveals
  inductive biases.
\newblock \emph{Psychonomic Bulletin \& Review}, 14(2):288--294.

\bibitem[{Kneser and Ney(1995)}]{kneser1995improved}
Reinhard Kneser and Hermann Ney. 1995.
\newblock Improved backing-off for m-gram language modeling.
\newblock In \emph{1995 international conference on acoustics, speech, and
  signal processing}, volume~1, pages 181--184. IEEE.

\bibitem[{Kuhn and Johnson(2013)}]{kuhn2013introduction}
Max Kuhn and Kjell Johnson. 2013.
\newblock An introduction to feature selection.
\newblock In \emph{Applied predictive modeling}, pages 487--519. Springer.

\bibitem[{Kuribayashi et~al.(2021)Kuribayashi, Oseki, Ito, Yoshida, Asahara,
  and Inui}]{kuribayashi2021lower}
Tatsuki Kuribayashi, Yohei Oseki, Takumi Ito, Ryo Yoshida, Masayuki Asahara,
  and Kentaro Inui. 2021.
\newblock \href {https://doi.org/10.18653/v1/2021.acl-long.405} {Lower
  perplexity is not always human-like}.
\newblock In \emph{Proceedings of the 59th Annual Meeting of the Association
  for Computational Linguistics}, pages 5203--5217.

\bibitem[{Langlois et~al.(2021)Langlois, Jacoby, Suchow, and
  Griffiths}]{langlois2021serial}
Thomas~A Langlois, Nori Jacoby, Jordan~W Suchow, and Thomas~L Griffiths. 2021.
\newblock Serial reproduction reveals the geometry of visuospatial
  representations.
\newblock \emph{Proceedings of the National Academy of Sciences}, 118(13).

\bibitem[{Lewis et~al.(2020)Lewis, Liu, Goyal, Ghazvininejad, Mohamed, Levy,
  Stoyanov, and Zettlemoyer}]{lewis2019bart}
Mike Lewis, Yinhan Liu, Naman Goyal, Marjan Ghazvininejad, Abdelrahman Mohamed,
  Omer Levy, Veselin Stoyanov, and Luke Zettlemoyer. 2020.
\newblock \href {https://doi.org/10.18653/v1/2020.acl-main.703} {{BART}:
  Denoising sequence-to-sequence pre-training for natural language generation,
  translation, and comprehension}.
\newblock In \emph{Proceedings of the 58th Annual Meeting of the Association
  for Computational Linguistics}, pages 7871--7880.

\bibitem[{Li et~al.(2020)Li, Roller, Kulikov, Welleck, Boureau, Cho, and
  Weston}]{li2019don}
Margaret Li, Stephen Roller, Ilia Kulikov, Sean Welleck, Y-Lan Boureau,
  Kyunghyun Cho, and Jason Weston. 2020.
\newblock \href {https://doi.org/10.18653/v1/2020.acl-main.428} {Don{'}t say
  that! making inconsistent dialogue unlikely with unlikelihood training}.
\newblock In \emph{Proceedings of the 58th Annual Meeting of the Association
  for Computational Linguistics}, pages 4715--4728.

\bibitem[{Linzen and Baroni(2021)}]{linzen2020syntactic}
Tal Linzen and Marco Baroni. 2021.
\newblock Syntactic structure from deep learning.
\newblock \emph{Annual Review of Linguistics}, 7(1).

\bibitem[{McAllester(2019)}]{bertconsistency}
David McAllester. 2019.
\newblock A consistency theorem for {BERT}.
\newblock Retrieved November 1, 2021 from
  {\url{https://machinethoughts.wordpress.com/2019/07/14/a-consistency-theorem-for-bert/}}.

\bibitem[{Meister and Cotterell(2021)}]{meister2021language}
Clara Meister and Ryan Cotterell. 2021.
\newblock \href {https://doi.org/10.18653/v1/2021.acl-long.414} {Language model
  evaluation beyond perplexity}.
\newblock In \emph{Proceedings of the 59th Annual Meeting of the ACL}, pages
  5328--5339.

\bibitem[{Meylan et~al.(2021)Meylan, Nair, and
  Griffiths}]{meylan2021evaluating}
Stephan~C Meylan, Sathvik Nair, and Thomas~L Griffiths. 2021.
\newblock Evaluating models of robust word recognition with serial
  reproduction.
\newblock \emph{Cognition}, 210:104553.

\bibitem[{Neville and Jensen(2007)}]{neville2007relational}
Jennifer Neville and David Jensen. 2007.
\newblock Relational dependency networks.
\newblock \emph{Journal of Machine Learning Research}, 8(3):653--692.

\bibitem[{Pillutla et~al.(2021)Pillutla, Swayamdipta, Zellers, Thickstun,
  Welleck, Choi, and Harchaoui}]{pillutla2021mauve}
Krishna Pillutla, Swabha Swayamdipta, Rowan Zellers, John Thickstun, Sean
  Welleck, Yejin Choi, and Zaid Harchaoui. 2021.
\newblock Mauve: Measuring the gap between neural text and human text using
  divergence frontiers.
\newblock \emph{Advances in Neural Information Processing Systems}, 34.

\bibitem[{Rafferty et~al.(2014)Rafferty, Griffiths, and
  Klein}]{rafferty2014analyzing}
Anna~N Rafferty, Thomas~L Griffiths, and Dan Klein. 2014.
\newblock Analyzing the rate at which languages lose the influence of a common
  ancestor.
\newblock \emph{Cognitive Science}, 38(7):1406--1431.

\bibitem[{Salazar et~al.(2020)Salazar, Liang, Nguyen, and
  Kirchhoff}]{salazar2019masked}
Julian Salazar, Davis Liang, Toan~Q. Nguyen, and Katrin Kirchhoff. 2020.
\newblock \href {https://doi.org/10.18653/v1/2020.acl-main.240} {Masked
  language model scoring}.
\newblock In \emph{Proceedings of the 58th Annual Meeting of the Association
  for Computational Linguistics}, pages 2699--2712.

\bibitem[{Sanborn et~al.(2010)Sanborn, Griffiths, and
  Shiffrin}]{sanborn2010uncovering}
Adam~N Sanborn, Thomas~L Griffiths, and Richard~M Shiffrin. 2010.
\newblock Uncovering mental representations with {Markov chain Monte Carlo}.
\newblock \emph{Cognitive psychology}, 60(2):63--106.

\bibitem[{Shane(2019)}]{shane2019you}
Janelle Shane. 2019.
\newblock \emph{You look like a thing and I love you}.
\newblock Hachette UK.

\bibitem[{Takahashi and Tanaka-Ishii(2017)}]{takahashi2017neural}
Shuntaro Takahashi and Kumiko Tanaka-Ishii. 2017.
\newblock Do neural nets learn statistical laws behind natural language?
\newblock \emph{PloS one}, 12(12):e0189326.

\bibitem[{Takahashi and Tanaka-Ishii(2019)}]{takahashi2019evaluating}
Shuntaro Takahashi and Kumiko Tanaka-Ishii. 2019.
\newblock Evaluating computational language models with scaling properties of
  natural language.
\newblock \emph{Computational Linguistics}, 45(3):481--513.

\bibitem[{Tenney et~al.(2019)Tenney, Das, and Pavlick}]{tenney2019bert}
Ian Tenney, Dipanjan Das, and Ellie Pavlick. 2019.
\newblock {BERT} rediscovers the classical {NLP} pipeline.
\newblock In \emph{Proceedings of ACL}, page 4593–4601.

\bibitem[{Toutanova et~al.(2003)Toutanova, Klein, Manning, and
  Singer}]{toutanova2003feature}
Kristina Toutanova, Dan Klein, Christopher~D Manning, and Yoram Singer. 2003.
\newblock Feature-rich part-of-speech tagging with a cyclic dependency network.
\newblock In \emph{Proceedings of the 2003 Human Language Technology Conference
  of the North American Chapter of the Association for Computational
  Linguistics}, pages 252--259.

\bibitem[{Wang and Cho(2019)}]{wang2019bert}
Alex Wang and Kyunghyun Cho. 2019.
\newblock {BERT} has a mouth, and it must speak: {BERT} as a {M}arkov {R}andom
  {F}ield language model.
\newblock In \emph{Proceedings of the Workshop on Methods for Optimizing and
  Evaluating Neural Language Generation (NeuralGen)}, page 30–36.

\bibitem[{Warstadt et~al.(2020)Warstadt, Parrish, Liu, Mohananey, Peng, Wang,
  and Bowman}]{warstadt2019blimp}
Alex Warstadt, Alicia Parrish, Haokun Liu, Anhad Mohananey, Wei Peng, Sheng-Fu
  Wang, and Samuel~R Bowman. 2020.
\newblock Blimp: The benchmark of linguistic minimal pairs for english.
\newblock \emph{Transactions of the Association for Computational Linguistics},
  8:377--392.

\bibitem[{Warstadt et~al.(2019)Warstadt, Singh, and
  Bowman}]{warstadt2019neural}
Alex Warstadt, Amanpreet Singh, and Samuel~R Bowman. 2019.
\newblock Neural network acceptability judgments.
\newblock \emph{Transactions of the Association for Computational Linguistics},
  7:625--641.

\bibitem[{Xu and Griffiths(2010)}]{xu2010rational}
Jing Xu and Thomas~L Griffiths. 2010.
\newblock A rational analysis of the effects of memory biases on serial
  reproduction.
\newblock \emph{Cognitive psychology}, 60(2):107--126.

\end{thebibliography}
\bibliographystyle{acl_natbib}

\renewcommand{\thefigure}{S\arabic{figure}}
\renewcommand{\thetable}{S\arabic{table}}
\setcounter{table}{0}
\setcounter{figure}{0}

\section*{Appendix A: Baseline details}

Wikipedia sentences were randomly selected from the full sentencized corpus English Wikipedia that tokenized to 12, 21, and 37 WordPiece tokens for the short, medium, and long conditions, respectively.
These sentences were also chosen to span a broad range of sentence probabilities under BERT (i.e. $\log P(p_1, \dots, p_n) = \sum_k \log P(p_k | p_{-k})$).

For our ngram baseline, we trained a 5-gram model with Kneser-Ney smoothing \cite{kneser1995improved} on English Wikipedia using the \texttt{kenlm} library \cite{heafield2011kenlm}, and generated sentences of length 10 by sampling from the resulting conditional distributions. 
Because this model stripped punctuation, and was therefore unable to emit an ``end of sentence'' token, we expected it to serve as a lower bound on the naturalness scale.

For our LSTM baseline, we used the network pre-trained by \citet{gulordava2018colorless} on English Wikipedia.
This model was trained to emit an end of sentence (\texttt{<eos>}) token, allowing us to rejection sample to obtain sentences that were exactly 10 words long with no unknown words (i.e. \texttt{<unk>} tokens).
Because it was not trained with a \texttt{<start>} token, however, we needed to initialize it with the initial word of the sentence. 
We randomly selected this initial word from a small set of common sentence openers (e.g. \texttt{the}, \texttt{a}, \texttt{it}, \texttt{his}, \texttt{her}).
As a result of our initial token selection, this model does not precisely sample from its true prior over sentences.
Thus, it is best viewed as another baseline of sentences rather than as a careful architectural comparison. 

Because we were asking participants to judge the naturalness of complete \emph{sentences}, we did not want to include samples which clearly violated sentencehood, as these would not be informative (e.g. fragments from Wikipedia that were incorrectly sentencized and ended with an abbreviation,  bibliographic text like ``korsakov (1976) r.s.,'' or table markdown with pipes like ``$|$ a $|$ b $|$'').
We automatically removed any sentences containing pipes or ending with colons or semicolons, as these were associated with sentencizer inconsistency, as well as sequences that contained multiple sentences (according to our sentencizer). 
Finally, the authors took a manual pass to exclude other non-sentential fragments from the stimulus set.

% latex table generated in R 4.0.3 by xtable 1.8-4 package
% Mon Nov 15 01:12:32 2021
\begin{table*}[hbt!]
\centering
\begin{tabular}{rlrrrr}
  \hline
 & term & estimate & std.error & statistic & p.value \\ 
  \hline
1 & (Intercept) & 67.33 & 1.14 & 59.08 & $<0.001$ \\ 
  2 & short vs. long (GSN) & -14.49 & 1.60 & -9.08 &$<0.001$ \\ 
  3 & short vs. medium (GSN) & -10.21 & 1.60 & -6.39 & $<0.001$ \\ 
  5 & GSN vs. LSTM (short) & -28.60 & 2.04 & -14.05 & $<0.001$ \\ 
  6 & GSN vs. MH (short) & -14.76 & 1.59 & -9.26 & $<0.001$ \\ 
  7 & GSN vs. ngram (short) & -54.26 & 2.00 & -27.07 & $<0.001$ \\ 
  8 & GSN vs. wiki (short) & 10.40 & 1.70 & 6.13 & $<0.001$ \\ 
  13 & interaction (short vs. long; GSN vs. MH) & -12.31 & 2.23 & -5.51 & $<0.001$ \\ 
  14 & interaction (short vs. medium; GSN vs. MH) & -7.33 & 2.23 & -3.29 & $<0.001$ \\ 
  17 & interaction (short vs. long; GSN vs. wiki) & 11.22 & 2.39 & 4.70 &$<0.001$ \\ 
  18 & interaction (short vs. medium; GSN vs. wiki) & 5.56 & 2.37 & 2.35 & 0.02 \\ 
   \hline
\end{tabular}
\caption{Fixed effect estimates for regression on human scores. Length class and source are dummy coded with short lengths and GSN as baselines.}
\label{tab:behaviorregression}
\end{table*}

\section*{Appendix B: Corpus details}
We downloaded cleaned Wikipedia data provided by GluonNLP (\href{https://github.com/dmlc/gluon-nlp/tree/master/scripts/datasets/pretrain_corpus}{https://github.com/dmlc/gluon-nlp/tree/master/scripts/datasets/pretrain\_corpus}), and BookCorpus data from HuggingFace Datasets (\href{https://huggingface.co/datasets/bookcorpus}{https://huggingface.co/datasets/bookcorpus}).

\begin{figure}[t]
\begin{center}
\includegraphics[width=0.9\linewidth]{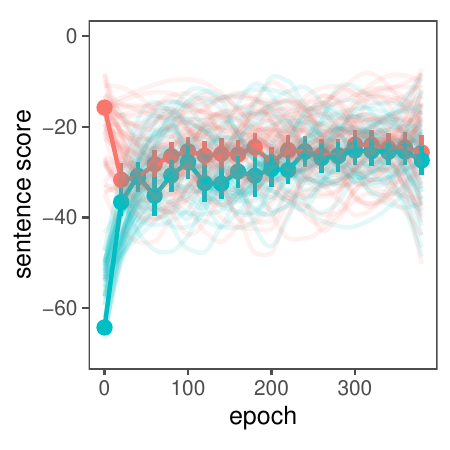}
\caption{We examine the convergence time by initializing different chains at different classes of sentences (red is high probability under BERT's energy function, blue is low probability). Faint lines show smoothed trajectories for individual chains and error bars are bootstrapped 95\% confidence intervals across chains.}
\label{fig:squeeze}
\end{center}
\end{figure}

\newpage

\begin{figure}[t!]
\begin{center}
\includegraphics[width=0.99\linewidth]{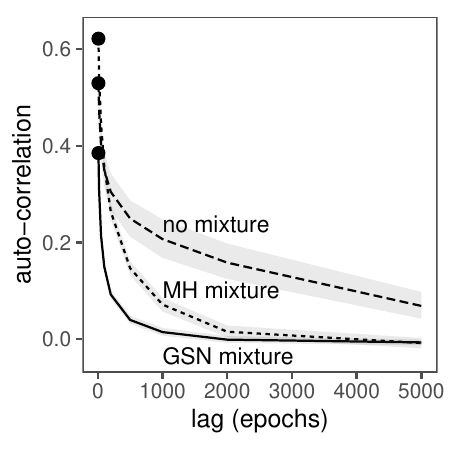}
\caption{MCMC methods like GSN and MH sampling tend to get stuck in local regions with high auto-correlation. We find that a minimal autocorrelation is achievable with lower lag (500 epochs between samples) using a mixture kernel with a constant probability of resetting the chain. Error ribbons are 95\% confidence intervals.}
\label{fig:ac}
\end{center}
\end{figure}

\begin{figure*}[ht]
\begin{center}
\includegraphics[width=0.99\linewidth]{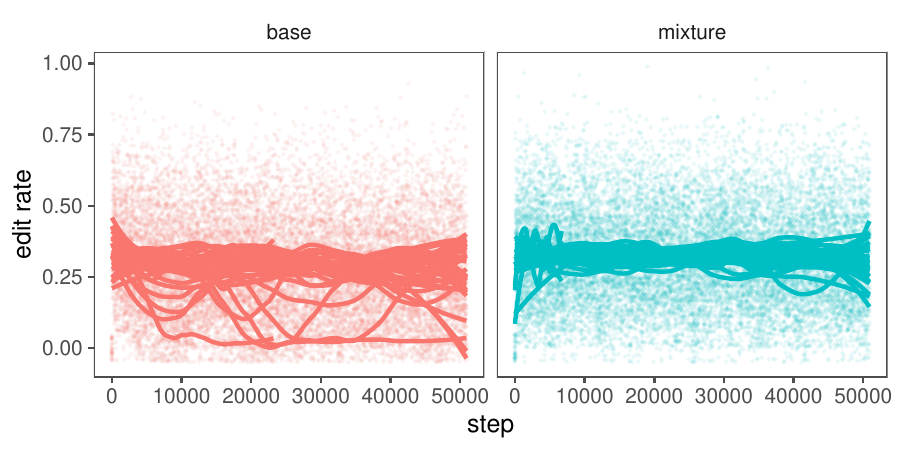}
\vspace{-1em}
\caption{Without mixing in a constant probability of returning to the initial distribution, the GSN chain (and MH chain, not shown) goes through periods of stasis with low edit rates (red curves), contributing to high autocorrelations.}
\vspace{-1em}
\label{fig:editrates}
\end{center}
\end{figure*}

\begin{figure*}[t!]
\begin{center}
\includegraphics[width=0.49\linewidth]{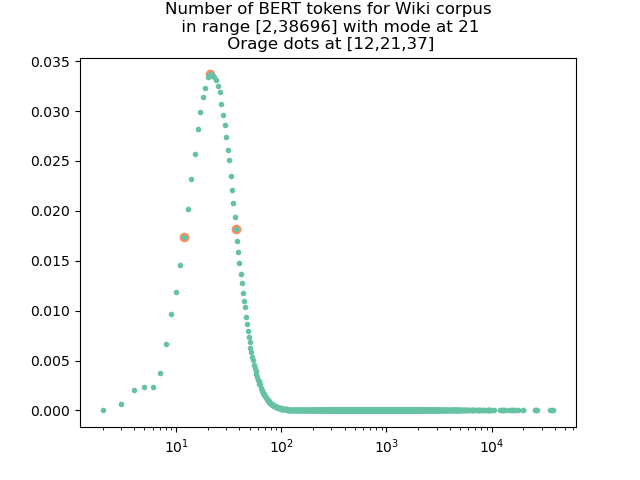}
\includegraphics[width=0.49\linewidth]{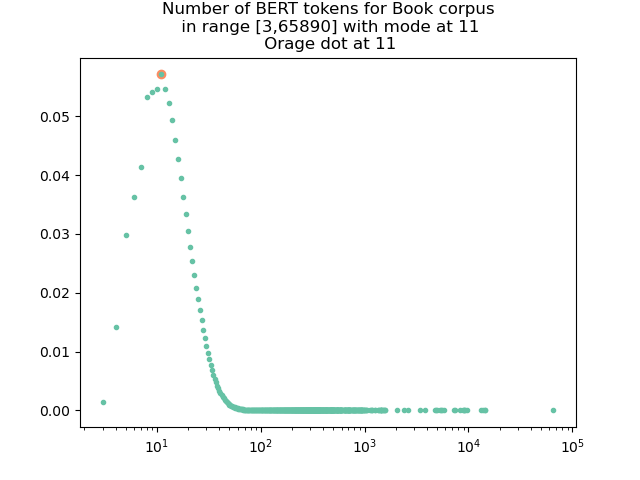}
\vspace{-1em}
\caption{Empirical distribution of sentence lengths in Wikipedia and BookCorpus training corpora, after WordPiece tokenization. For our corpus comparisons, we selected the modal Wikipedia sentence length of 21 tokens and the modal BookCorpus length of 11 tokens. For our human judgment experiment, we included baseline sentences only from Wikipedia for shorter (12 tokens) and longer sentences (37 tokens), with roughly equal prevalence in the corpus (orange dots).}
\vspace{-1em}
\label{fig:wikisentlength}
\end{center}
\end{figure*}

\begin{figure}[tbh]
\begin{center}
\includegraphics[width=0.75\linewidth]{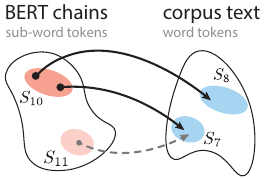}
\caption{There is a misalignment between the space of sentences obtainable by a BERT chain of a fixed token length (in sub-word tokens) and natural language sentences of a fixed length (in words). We consider the distribution of corpus sentences that are obtainable from a fixed-length BERT chain, which may decode to different lengths in natural text (black arrows).}
\label{fig:misalignment}
\end{center}
\end{figure}

\begin{figure}[ht]
\begin{center}
\includegraphics[width=.99\linewidth]{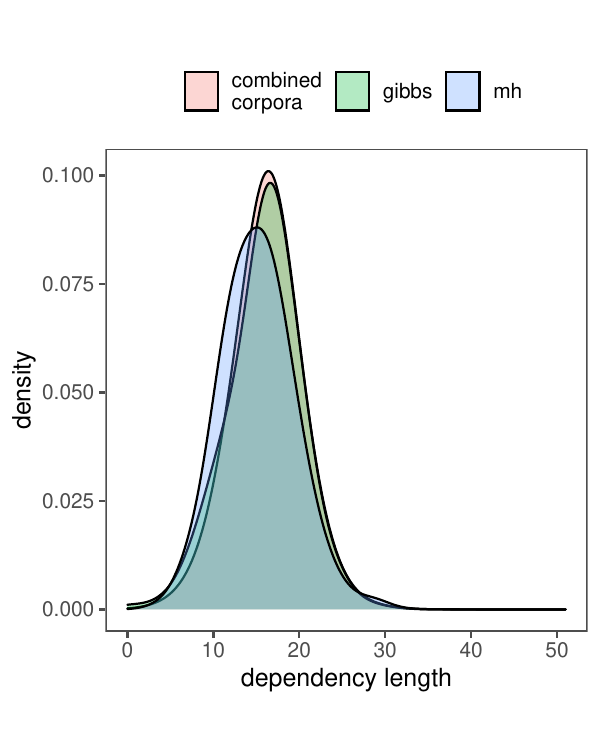}
\vspace{-1em}
\caption{Dependency distances are similar for sentences sampled from BERT's prior and sentences from its training corpus, but the BERT distribution is more bimodal and tends to skew simpler.}
\vspace{-1em}
\label{fig:dependency_distance}
\end{center}
\end{figure}

\begin{figure*}[t!]
\begin{center}
\includegraphics[width=0.9\linewidth]{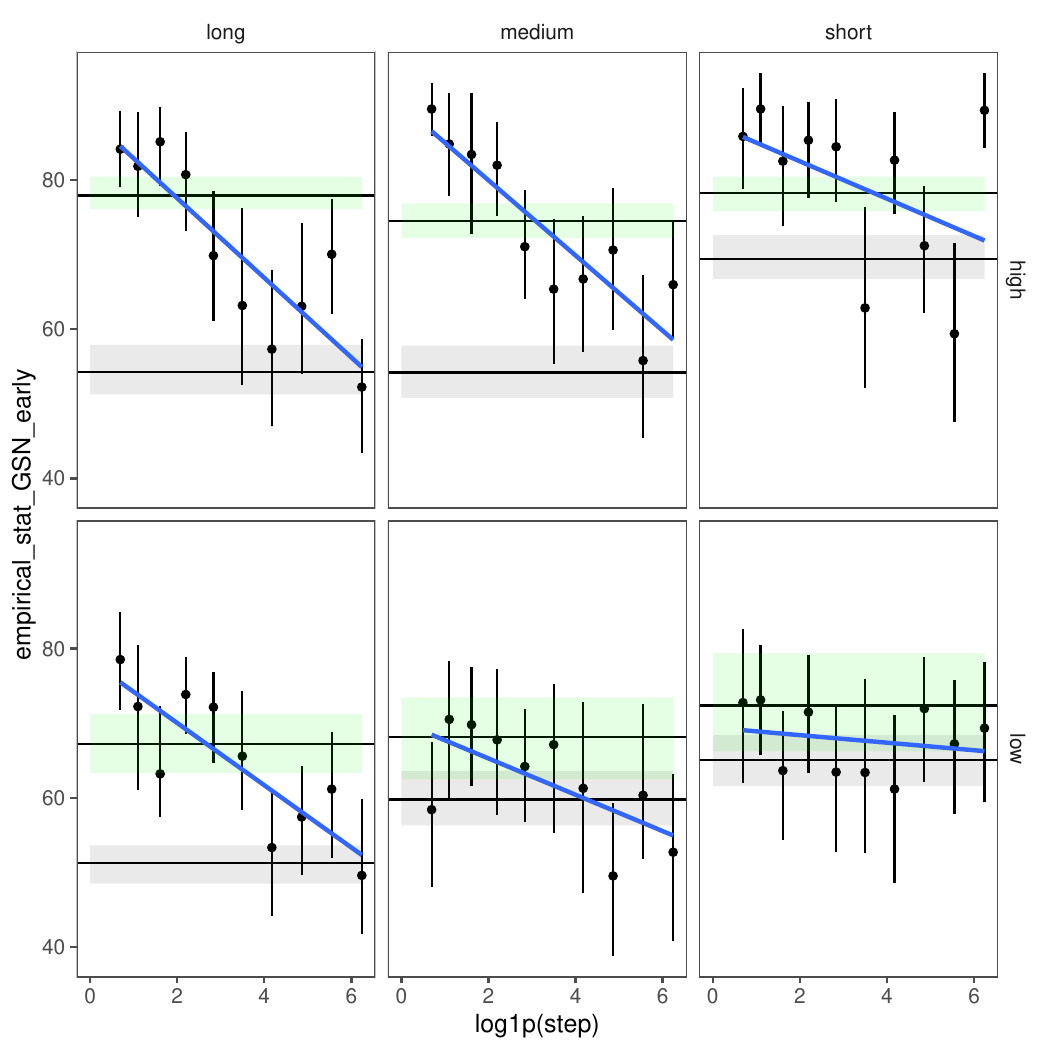}
\caption{Sentences gradually drift away from the initial distribution across the burn-in period. Light green region represents the 95\% confidence interval for the mean naturalness of Wikipedia sentences while grey region represents the same interval around the stationary distribution of the converged chain. Top row represents chains that are initialized at high-probability states, while bottom row is initialized in low-probability states.}
\label{fig:burnin}
\end{center}
\end{figure*}

% \begin{figure*}[ht]
% \begin{center}
% \includegraphics[width=0.99\linewidth]{figs/burn-in-itemlevel.pdf}
% \vspace{-1em}
% \caption{Variability in ratings across sentences for burn-in period. Each point represents the mean rating for a single sentence at the given step in the chain. Smoothing line uses a non-parametric loess fit.}
% \vspace{-1em}
% \label{fig:burnin_itemlevel}
% \end{center}
% \end{figure*}

% \begin{figure*}[ht]
% \begin{center}
% \includegraphics[width=0.6\linewidth]{figs/lexical-distributions_mh.pdf}
% \vspace{-1em}
% \caption{The lexical distribution of the sentences produced using Metropolis-Hastings algorithm is also calibrated to the corpus with a slightly lower correlation (r=0.64, among words that appeared more than 10 times).}
% \label{fig:lex_mh}
% \end{center}
% \end{figure*}

% \begin{figure*}[ht]
% \begin{center}
% \includegraphics[width=0.99\linewidth]{figs/prob-response-relationship.pdf}
% \vspace{-1em}
% \caption{Sentences that are scored by BERT as higher probability also tend to be rated as more natural by human speakers. Spearman correlations at the sentence level are significant, $p < 0.01$, for all three classes of BERT sentences, but not for Wikipedia sentences ($p = 0.39, p = 0.31, p = 0.24$, respectively), primarily due to the number of annotated examples. We removed a single `leverage point' for the long Wikipedia sentences with a log probability of -200, which moderately increased the correlation to $r = 0.17$ but did not influence the overall pattern of significance.}
% \vspace{-1em}
% \label{fig:probResponse}
% \end{center}
% \end{figure*}

\begin{table*}[ht]
    \centering
    %\footnotesize
    \begin{tabular}{c|c|p{250pt}}
      \multicolumn{2}{c|}{types of unnaturalness} & examples \\
      \hline\hline
      \multicolumn{2}{c|}{character-level} & He preened on a \begin{CJK}{UTF8}{min}
      レ
        \end{CJK} drink of copper. \\
      \hline
      \multirow{3}{*}{phrase-level} & \multirow{3}{*}{semantic} & The little wattled songbird, also called the Chink Warbler, Orange Garver or Quickcumber is a {\bf socially luscious} and habituated bird species.\\
      \hline
      \multirow{9}{*}{sentence-level} & \multirow{1}{*}{construction} & There were two hours before he {\bf made the walk}.\\
      \cline{2-3}
      & \multirow{1}{*}{out-of-context word} & No need to focus on {\bf bicycling}.\\
      \cline{2-3}
      & \multirow{2}{*}{self-contradictory} & The symbols $(\cdot \cdot \cdot)$ read as  $(\cdot \cdot \cdot)$  and  $(\cdot)$  are written as  $(\cdot \cdot \cdot\cdot\cdot)$ , not as $(\cdot)$.\\
      \cline{2-3}
     &  \multirow{5}{*} {repetition}& The college of arts and sciences, adjacent to the business school, is majoring in business.\\
      \cline{3-3}
      && He saw Cronus and Cronus, Cronus and James Cronus he saw Cronus and Cronus and Cronus and Cronus Cronus when he saw Cronus.
    \end{tabular}
    \caption{More examples of sentences from BERT's prior with low naturalness ratings.}
    \label{tab:more_examples}
\end{table*}

\end{document}